\documentclass{article}

\usepackage{arxiv}

\usepackage[utf8]{inputenc} 
\usepackage[T1]{fontenc}    
\usepackage{hyperref}       
\usepackage{url}            
\usepackage{booktabs}       
\usepackage{amsfonts}       
\usepackage{nicefrac}       
\usepackage{microtype}      
\usepackage{lipsum}
\usepackage{fancyhdr}       
\usepackage{graphicx}       
\graphicspath{{media/}}     
\usepackage{mdframed}
\usepackage{xcolor}
\usepackage{url}
\usepackage{amsmath}

\let\bld\boldsymbol

\title{Scalable Artificial Intelligence for Science:\\Perspectives, Methods and Exemplars
}

\author{
  Wesley Brewer, Aditya Kashi, Sajal Dash, Aristeidis Tsaris, Junqi Yin, Mallikarjun Shankar, Feiyi Wang \\
  National Center for Computational Sciences \\
  Oak Ridge National Laboratory \\
  Oak Ridge, TN USA 37380 \\
  \texttt{\{brewerwh,kashia,dashs,tsarisa,yinj,shankarm,fwang2\}@ornl.gov} \\
}

\begin{document}
\maketitle

\begin{abstract}
In a post-ChatGPT world, this paper explores the potential of leveraging scalable artificial intelligence for scientific discovery. We propose that scaling up artificial intelligence on high-performance computing platforms is essential to address such complex problems. This perspective focuses on scientific use cases like cognitive simulations, large language models for scientific inquiry, medical image analysis, and physics-informed approaches. The study outlines the methodologies needed to address such challenges at scale on supercomputers or the cloud and provides exemplars of such approaches applied to solve a variety of scientific problems.
\end{abstract}


\section{Main}

In light of ChatGPT's growing popularity, the transformative potential of AI in science becomes increasingly evident. 
Although a number of recent articles highlight the transformative power of AI in science \cite{schmidt2023ai,economist2023ai,wang2023scientific}, few provide specifics how to implement such methods at scale on supercomputers.
Using ChatGPT as an archetype, we argue that the success of such complex AI models results from two primary advancements: (1) the development of the transformer architecture, (2) the ability to train on vast amounts of internet-scale data. This process represents a broader trend within the field of AI where combining massive amounts of training data with large-scale computational resources becomes the foundation of scientific breakthroughs.

Several examples underscore the integral role of using large-scale computational resources and colossal amounts of data to achieve scientific breakthroughs. For instance, Khan et al. \cite{khan2022ai} used AI and large-scale computing for advanced models of black hole mergers, leveraging a dataset of 14 million waveforms on the Summit supercomputer. Riley et al. \cite{riley2023effect} made significant progress towards the understanding the physics of stratified fluid turbulence by being able to model the Prandtl number of seven, which represents ocean water at $20^\circ$ Celsius. Such simulations required being simulated using four trillion grid points, which required petabytes of storage \cite{couchman2023mixing}. The data are later analyzed using unsupervised machine learning techniques to understand the underlying mechanisms driving the turbulence \cite{de2019unsupervised}. Finally, the AlphaFold2 breakthrough in protein structure prediction became a reality due to a comprehensive database of 170,000 protein structures and an 11-day training period on 128 TPUv3 cores \cite{jones2022impact,jumper2021highly}. To further highlight this point, we note that Gordon Bell Prize winners have increasingly been AI-driven workflows at large scale -- each of which has revealed significant scientific insights \cite{zvyagin2022genslms,jia2020pushing,casalino2021ai,kurth2018exascale}.

As we move forward, we will delve deeper into these critical aspects, drawing from recent advances and discussing the potential future challenges and solutions in achieving scalability and performance in training large AI models. We examine methods to scale AI systems to the extreme and discuss future trends, while also shedding light on key perspectives shaping our view of scalable AI in science. Throughout this paper, we will elucidate several pivotal perspectives that shape our understanding of scalable AI in scientific applications. Specifically, we focus on:

\begin{enumerate}
    \item The indispensability of high-performance computing for achieving scientific breakthroughs, emphasizing that distinct problems may necessitate varying levels of computational scale.
    \item The distinct requirements that differentiate AI for Science (AI4S) from consumer-centric AI.
    \item The need for specific methods and architectures to accommodate scientific data as well as enforce laws of science. 
    \item The strategic shift from a singular, large monolithic model towards a synergistic mixture of experts (MoE) models.
    \item The imperative of scaling not merely in computational terms, but also across infrastructural resources and integrated research infrastructures (IRI).
\end{enumerate}

Our paper is laid out as follows: Section \ref{sec:gpt} discusses lessons that were learned from scaling GPT-3; Section \ref{sec:different} contrasts how scientific AI differs from consumer-grade AI; Section \ref{sec:sml} introduces specific types of techniques that are unique to scientific AI, such as soft-penalty constraints and neural operators; Section \ref{sec:scaling} discusses how to train and deploy neural networks at scale on supercomputers; Section \ref{sec:workflows} outlines the different types of AI-HPC workflows that are typically used in practice; and Section \ref{sec:conclusion} concludes with a summary as well as numerous perspectives on where we believe the field is headed. Along the way, we disperse various exemplars of using such methods, touching on various topics that employ different methods and data modalities, such as: LLMs for drug discovery; physics-informed approaches for modeling turbulence; medical document analysis for critical insights into cancer diagnosis and management; whole slide image analysis for cancer detection; and computational steering for neutron scattering experiments. 

\section{Lessons from GPT-3}
\label{sec:gpt}

Large models and data, whether from neural architecture or training size, require distribution across multiple GPUs for scalability, necessitating parallel scaling. Consider that it would take 288 years to train the GPT-3 large language model on a single NVIDIA V100 GPU \cite{liopenai}; however, using parallel scaling techniques, this time drops to 36 years on eight GPUs or seven months on 512 GPUs \cite{narayanan2021efficient}, and just 34 days using 1024 A100 GPUs \cite{narayanan2021efficient}. Besides faster training, scaling enhances model performance \cite{brown2020language}. 

Parallel scaling has two main approaches: model-based and data-based parallelism. \emph{Model-based} parallelism is needed when models exceed GPU memory capacity, like GPT-3's 800GB size versus an NVIDIA A100's 80GB memory \cite{brown2020language}. \emph{Data-based} parallelism arises from the large amounts of data that are required to train such models, e.g., GPT-3 requires 45TB of text data for training \cite{neuwirth2021parallel}. We explore this subject in more detail in Section \ref{sec:sml}. 

\begin{figure}[t] 
\begin{mdframed}[backgroundcolor=black!10,rightline=false,leftline=false]
\textbf{Example: LLMs for Drug Discovery}

In 2017 Transformer-based Language Models (LLMs) have marked a significant AI breakthrough in scientific research \cite{vaswani2017attention}. One exciting and promising application of these LLMs is in the realm of drug discovery \cite{andrew2022}. Although current chemical databases provide access to billions of molecules, they only represent a small fraction of the vast space of potentially synthesizable molecules. However, by employing tokenization and mask prediction techniques, LLMs can utilize extensive datasets to autonomously learn recurring sub-sequences (known as structural components) and potential rearrangements, thereby enabling efficient exploration of chemical space. This innovative approach follows the foundation model paradigm, where unsupervised pre-training on large datasets is combined with fine-tuning on small labeled datasets for specific downstream tasks. By integrating pre-trained models for molecule and protein sequences, fine-tuning leverages data from prior experiments, enhancing performance and prediction.
\end{mdframed}
\end{figure}

\section{How AI for Science is Different}
\label{sec:different}

Given that ``data is the lifeblood of AI'', it is important to understand how scientific data in AI4S differs from consumer or commercial data and the techniques to handle it. Scientific data is often more sparse and less accessible than commercial data, typically provided via expensive experiments or simulations. This data might have fewer labels or show asymmetry, with some samples labeled and others not.

For scientific model validation, high-precision floating-point numbers are common, while consumer models often inference as 8-bit integers. The requirements for AI in engineering, like self-driving cars or virtual sensors in helicopters, are much stricter than for photo generation or modification. Trustworthiness is paramount for AI4S applications \cite{matsuoka2023deployment}. 

Moreover, neural networks in physics-based applications might need to impose boundary conditions or conservation laws. This is particularly true for surrogate models, which replace parts of larger simulations, like machine-learned turbulence models. 

\section{Methods for Scientific Machine Learning}
\label{sec:sml}

Unlike conventional machine learning in consumer IT and commerce, natural sciences present unique requirements and opportunities. Three other pillars of investigation—theory, experiment, and numerical computation—have long provided valuable insights. Exploring the dynamic interplay between these pillars and machine learning is of great interest. Certain natural science fields demand greater accuracy and reliability than typical machine learning. Aerospace engineering, nuclear physics and engineering, and medical research are prime examples. Furthermore, limited data availability poses challenges, particularly in fields such as medical research with privacy concerns and costly experiments.

This section describes methods of scientific machine learning, which broadly cater to the following needs of scientific computing:
\begin{itemize}
    \item Utilizing known domain knowledge, particularly physics; this is commonly in the form of partial differential equations (PDEs)
    \item Dealing with multiple inputs and outputs which are not independent numbers but represent functions over space and time
    \item Identifying patterns and dynamics in terms of mathematical formulae
    \item Error estimates and uncertainty quantification
\end{itemize}
These factors have led to the emergence of methods that we collectively refer to as ``scientific machine learning''.

Many of these methods deal with differential equations, particularly PDEs. The PDE stands as a far-reaching, expressive, and ubiquitous modeling tool in mathematics and the sciences.
A PDE model includes the equation, spatial domain, boundary conditions and initial conditions. It involves sets of functions (the `state') defined in the domain of interest in space-time. Solving PDEs is crucial in prediction and analysis (forward problem) and design, control, and decision-making (backward problem). While PDEs are potent tools, relying solely on them may lead to challenges when dealing with complex phenomena. In computational fluid dynamics, for example, obtaining highly accurate solutions using direct numerical simulation can be prohibitively expensive for realistic turbulent flows, while approximating them with turbulence models can lead to incorrect predictions. Augmenting PDEs with data-driven models has been explored to address these issues and improve model accuracy and cost. Efforts to integrate physics-based and machine learning models have yielded promising results.

\subsection{Soft Penalty Constraints}

\begin{figure}
    \centering
    \includegraphics[width=8cm]{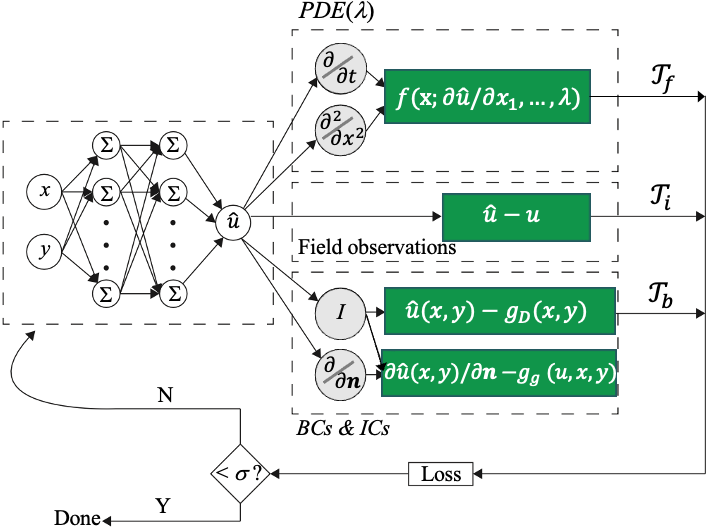}
    \caption{PINN architecture from Chen et al. \cite{chen_physics-informed_nanooptics_2020}}
    \label{fig:pinn}
\end{figure}

Physics-informed neural networks (PINNs) are an increasingly common approach for modeling PDEs \cite{raissi_physics_informed_2019}. These networks take spatial coordinates $(x,y,z,t)$ as input and output an approximate solution $\bld{u}(x,y,z,t)$ to the PDE $\bld{R}(\bld{u}) = \bld{0}$ at that location and time (see figure \ref{fig:pinn}). Incorporating the PDE residual norm $\lVert \bld{R}(\bld{u})\rVert$ in the loss function ensures that the model optimizer minimizes both the data loss and the PDE residual, leading to a satisfying approximation of the physics. Raissi et al. \cite{raissi_physics_informed_2019} use simple feed-forward neural networks with a $\tanh$ activation function for a 1D geometry, while Chen et al. \cite{chen_physics-informed_nanooptics_2020} apply PINNs to solve 2D inverse problems in nano-optics, demonstrating the promising role of AI in solving such challenges.

\subsection{Neural Ordinary Differential Equations (NODE)} 

The idea behind the original NODE \cite{chen2018neural} landmark paper is to use a neural network to learn the time derivative of a dataset, and then use an ODE solver to time integrate the neural network. The difficulty of such approaches lies in how to perform backpropagation through the ODE solver, which is handled using the adjoint sensitivity approach. Such methods are useful to solve temporal forecasting problems, due to the fact that they perform well in extrapolation mode. There have been a numerous variants of NODE since the original paper, including: Augmented (ANODE) \cite{dupont2019augmented}, Stochastic (SDE-NODE) \cite{tzen2019neural}, Discretely updated (DUNODE) \cite{li2021discretely}, Higher-order (HONODE) \cite{massaroli2020dissecting}, Controlled (CNODE) \cite{poli2020controlled}, and Piecewise (PNODE) \cite{chen2021piecewise} variants.

\subsection{Universal Differential Equations} 

Rackauckas et al. \cite{rackauckas_ude_2021} introduced `universal differential equations' (UDE) where neural networks model components of the equations. UDEs are particularly useful when certain physics terms are well-defined using operators, while others require data-driven modeling. These equations can be stochastic, and some components may be algebraic, leading to differential algebraic equation (DAE) systems. The authors provide Julia-based software toolkits for working with UDEs. They demonstrate that training neural network components of UDEs and employing sparse regression leads to improved equation discovery compared to direct sparse regression on the entire differential equation (e.g., SINDy \cite{sindy}). UDEs can also encompass PDEs. For example, they demonstrate learning one-dimensional reaction-diffusion equations by using a convolutional neural network (CNN) for the diffusion operator and a simple feedforward network for the nonlinear reaction operator, trained for specific geometries and fixed boundary conditions. Moreover, the authors showcase learning a sub-model for incompressible Navier-Stokes equations, the averaged temperature flux model, employing the Boussinesq approximation for gravity.

\subsection{Neural Operators}

An interesting and important area of research involves `neural operators' — maps between function spaces using neural networks. These operators are especially useful for PDEs and scenarios dealing with functions or signals, like non-uniform material properties. For instance, thermal conductivity $\nu$ may vary in space in the heat equation  
\begin{equation}
 \frac{\partial u}{\partial t} - \nu(\bld{x}) \nabla^2 u = 0.
 \label{eqn:pde_heat}
\end{equation}
The objective is now to map any given input function $\nu$ to the solution $u$ of the PDE, unlike regular models (like most PINNs) that take the coordinates $(\bld{x},t)$ as input and output the \emph{value} of $u$ at that point.

\begin{figure}
    \centering
    \includegraphics[width=8cm]{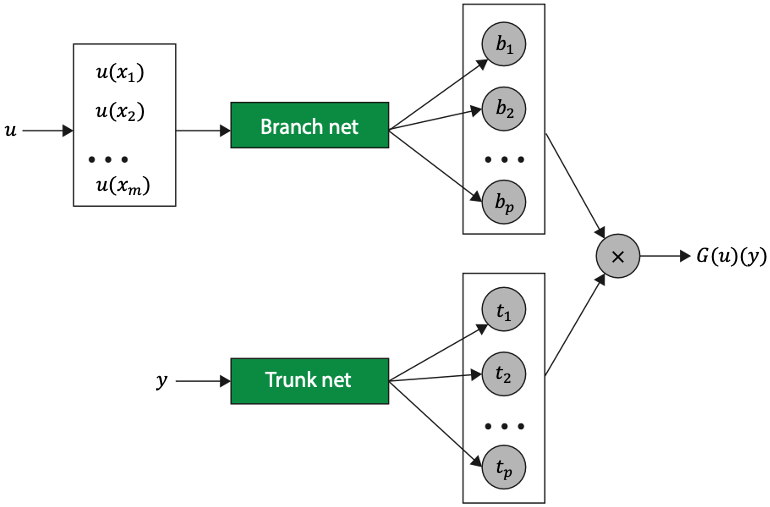}
    \caption{DeepONet architecture \cite{deeponet_2021}.}
    \label{fig:deeponet}
\end{figure}

Lu et al. \cite{deeponet_2021} introduced `DeepONet', a deep operator network, that learns linear or nonlinear operators. They utilize branch and trunk networks (figure \ref{fig:deeponet}), demonstrating superior performance over other neural networks. Li et al. \cite{li_neural_operator_2020} introduced `graph kernel networks' for solving elliptic PDEs, using graph neural networks to learn kernel functions. An innovative new technique that leverages classical Fourier analysis is the Fourier neural operator \cite{li_fourier_op_2021, pathak2022fourcastnet}, though it also has geometric limitations.
Kurth et al. \cite{pathak2022fourcastnet} trained their neural operator model on up to 3808 GPUs on some of the world's largest supercomputing systems. Their adaptive Fourier neural operator architecture obtains reasonably good results with high, scalable performance.
Neural operators can also incorporate known physics \cite{goswami_physics-informed-op_2022, li_physics-informed-op_2022}.

\subsection{Coarse-graining}
Machine learning can be used to improve simulations by coarse-graining using data-driven discretization \cite{bar2019learning}. Data-driven discretization allows for the high fidelity solution of PDEs on much coarser grids. Such methods were inspired by the concept of super-resolution of images using a generative adversarial network (GAN) \cite{ledig2017photo}.

\subsection{Temporal forecasting} 

Many scientific problems are transient in nature, such as fluid dynamics in a pumping heart. Lim and Zohren \cite{lim2021time} suggest that three particular neural network architectures are effective for temporal forecasting type problems: (1) CNN with dilated causal convolutional layers -- sometimes referred to as Temporal CNNs (TCNN), (2) recurrent neural networks (RNN) such as long short-term memory (LSTM), and (3) attention-based models such as the transformer architecture \cite{vaswani2017attention}. One point to note for such problems is that the training data must be first converted to time sequences, given a window size. 

One novel technique for modeling temporal forecasting is neural  ordinary differential equations (NODE) \cite{chen2018neural}, where a neural network is used to learn the time derivative of a dataset, and then an ODE solver is used to integrate the neural network in time. The difficulty of such approaches lies in how to perform backpropagation through the ODE solver, which is handled using the adjoint sensitivity approach. Such methods are useful to solve temporal forecasting problems, e.g., turbulence forecasting \cite{portwood2019turbulence}, especially due to the fact that they perform well in extrapolation mode. There have been a numerous variants of NODE since the original paper, which are covered in \cite{zhang2022pnode}. 

\subsection{Symbolic Regression}

Symbolic regression is a method for learning explicit mathematical representations from data. One popular method for this can be traced back to genetic programming by \cite{koza1992genetic}. Recent methods have leveraged machine learning techniques, e.g., Cranmer et al. \cite{cranmer2020discovering}, and developed scalable implementations of such methods, e.g., Biggio et al. \cite{biggio2021neural}.
There are both linear \cite{sindy} and nonlinear methods \cite{landajuela2022unified}, though both classes of methods can learn nonlinear functional forms.

\subsection{Uncertainty Quantification} 

A challenge in using AI for science is the lack of reliability estimates for predictions. Scientific models should not only predict outcomes but also quantify uncertainty \cite{psaros2023uncertainty}. Methods such as Gaussian Process Regression (GPR) are in essence neural networks that predict probability distributions. GPFlow \cite{matthews2017scalable} is a framework built on TensorFlow Probability that is able to solve such problems at scale.  Physics-informed Generative Adversarial Networks (PI-GAN) have also been able to predict variables along with uncertainty \cite{yang2019adversarial}. 
MonteCarlo Dropout \cite{gal2016dropout} is another technique which uses the idea of dropout -- a method often used for regularization in the training of neural networks \cite{srivastava2014dropout} -- but instead of using it during training, it is used during inference. The state-of-the-art for uncertainty prediction are prediction interval methods, such as PI3NN \cite{liu2021pi3nn}, which uses a linear combination of three neural networks to predict confidence levels. The PI3NN method is also able to predict when unseen samples lie outside the bounds of the training data, i.e., out-of-distribution (OOD). 

\begin{figure}[t] 
\begin{mdframed}[backgroundcolor=black!10,rightline=false,leftline=false]
\textbf{Example: Modeling Turbulence}

Richard Feynman once described turbulence as ``the most important unsolved problem of classical physics''. There have been numerous research efforts towards the efforts of using AI/ML to understand and model turbulence. These generally fall into either \textit{wall-bounded} turbulence, such as found in engineering problems of airplanes or ships, or \textit{wall-free} turbulence, which is found in geophysical flows, such as found in atmospheric and oceanic flows. Wall-bounded flows generally train on Direct Numerical Simulation data to train a model which is deployed using strategies shown in Fig. \ref{fig:yin} -- a recent survey of such methods may be found in \cite{bhushan2023assessment}. On the other hand, state-of-the-art approaches for modeling atmospheric turbulence are using NODE-type techniques, such as described by Shankar et al. \cite{shankar2022validation}. Finally, there is also ongoing work which uses unsupervised learning to analyze the structure of stratified turbulent flows, such as the work by Couchman et al. \cite{couchman2023routes}.
\end{mdframed}
\end{figure}

\subsection{Surrogate Models} 

\begin{figure}
    \centering
    \includegraphics[width=10cm]{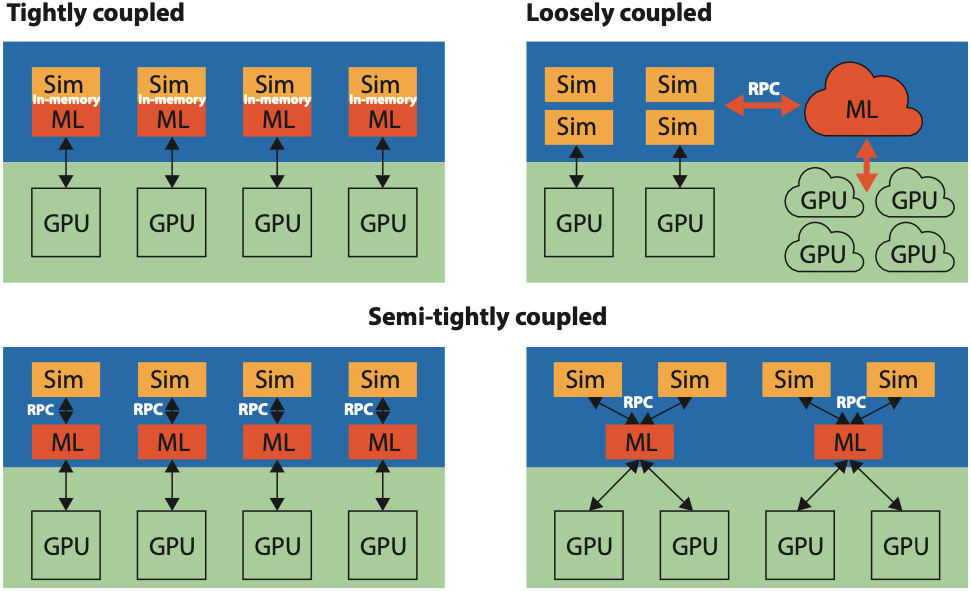}
    \caption{Different strategies for deploying machine-learned surrogate models on HPC \cite{yin2022strategies}.}
    \label{fig:yin}
\end{figure}

Beyond efforts to replace traditional simulations with AI/ML methods like NVIDIA Modulus \cite{hennigh2021nvidia}, there are increasing efforts on hybrid AI-simulations, also known as cognitive simulations (CogSim). A common example of this type of workflow is machine-learned \textit{surrogate modeling}, in which case only a portion of the overall workflow is replaced with a machine-learned model. For example, rather than use a traditional turbulence model, such as $k$-$\omega$, data-driven machine-learned turbulence models can be learned, which can often outperform empirically based models. In this case either the turbulence production term can be modeled directly, or a predictor-corrector type implementation can be employed, where the machine-learned model corrects the prediction of the traditional model \cite{zhang2022predictor, bhushan2023assessment}. Such methods may be deployed using either in-memory or remote inference strategies via remote procedure calls (RPC) as shown in Fig. \ref{fig:yin}. Partee et al. \cite{partee2022using} present SmartSim, a framework for augmenting simulations to inference machine-learned surrogate models at scale. Brewer et al. \cite{brewer2021production} study inferencing at petascale for surrogate models used in rotorcraft aerodynamics with both a RedisAI-based framework as well as TensorFlow Serving \cite{olston2017tensorflow}. Boyer et al. \cite{boyer2022scalable} extend this study to fully integrate inference serving techniques in to C++ computational physics simulations in a scalable way using MPI.

Finally, one of the more interesting implementations of blending simulations and machine-learned models is the work by Vlachas et al. \cite{vlachas2022multiscale}, in which they define ``algorithmic alloys'' to meld machine learning approaches with more traditional approaches to learn effective dynamics of complex systems. 

A new area of implementation that also blends simulation and machine-learned models is the area of \textit{digital twins} \cite{kapteyn2021mathematical}, where there is not only simulation and machine-learned models intercommunicating, but also real-time telemetry data from the physical twin being assimilated into the simulation. 

\subsection{Deployment and Distributed Inference} 

Inference may be performed using either an in-memory approach or an remote approach (client-server architecture) as shown in Fig. \ref{fig:yin}. In the case of remote inference using remote procedure calls (RPC), it is typically scaled \textit{asynchronously}, where multiple inference servers run in parallel, and inference requests are sent in an embarrassingly parallel fashion. There are generally two ways scaling is achieved: (a) using a load balancer (e.g., haproxy), or (b) message-passing approach using the message passing interface (MPI). While the latter method can be more performant than the former, the former approach may be better for handling node faults, and thus may be more appropriate for high availability.

 Yin et al. \cite{yin2022strategies} applied such techniques for thermodynamics surrogate models, and studied different types of coupling -- from strong to weak -- and demonstrate scaling up to 1000 GPUs. 

LLM on the other hand must be deployed with \textit{synchronous} inference, where the model must be distributed across multiple accelerators. Pope et al. \cite{pope2022efficiently} studied inference performance of Transformer models with 500B+ parameters distributed across 64 TPU v4 chips. 

\begin{figure}[th] 
\begin{mdframed}
[backgroundcolor=black!10,rightline=false, leftline=false]
\textbf{Example: Medical Document Analysis}

Analysis of pathology reports provides critical insight into cancer diagnosis and management. A number of cancer characteristics are coded manually from the cancer pathology report as part of Surveillance, Epidemiology,
and End Results (SEER) database. A recent collaborative effort between DOE and NCI has tried to build AI-based models for pathology information extraction tasks namely: site, subsite, laterality, histology, and behavior.
In that work, we build a transformer model that can effectively accommodate the length of typical cancer pathology reports. We use 2.7 million pathology reports from six SEER cancer registries to train purpose-built sparse transformer models such as pathology BigBird model \cite{zaheer2020big}. BigBird
model is a sparse attention-based transformer model built for long documents compared to popular dense attention-based models such as BERT.
\end{mdframed}
\end{figure}

\section{Methods of Parallel Scaling}
\label{sec:scaling}
 
\begin{figure}[p]
    \centering
    \includegraphics[width=10cm]{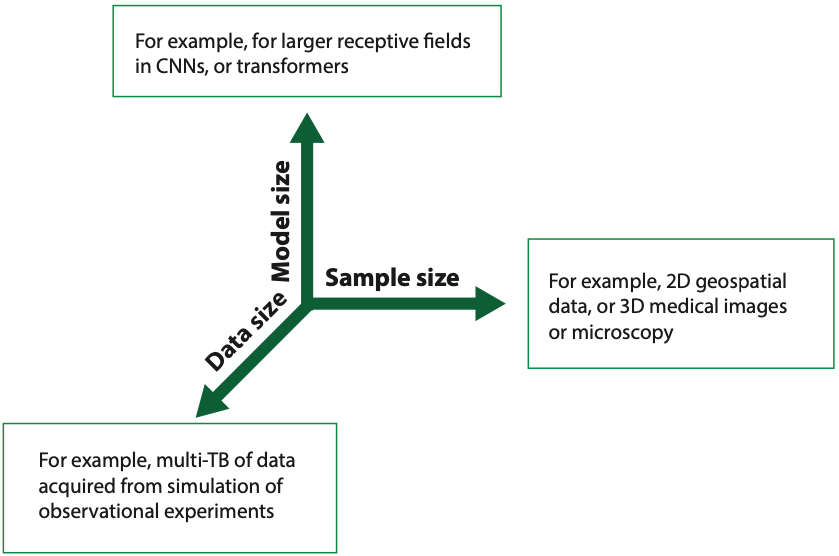}
    \caption{Showing the orthogonal nature of scalable AI workflows.}
    \label{fig:parallel_sketch}
\end{figure}

Many frameworks have been engineered to accommodate the increasing demand for highly scalable AI systems and complex workflows. Examples of such frameworks include: Horovod \cite{sergeev2018horovod}, Megatron-LM \cite{narayanan2021efficient}, DeepSpeed \cite{rasley2020deepspeed}, and Fully Sharded Data Parallel (FSDP) \cite{zhao2023pytorch}. Each of these frameworks employs different strategies to enable scalability, primarily falling into two categories of parallelism techniques: Data Parallelism and Model Parallelism, with Pipeline Parallelism being the most prominent implementation of the latter. Intriguingly, most of these approaches are orthogonal to each other, as illustrated in Fig. \ref{fig:parallel_sketch}, meaning that they address unique challenges and bottlenecks in system performance. By understanding these nuances, researchers can more effectively leverage the right frameworks and parallelism strategies to optimize their AI deployments. For example, recent studies have investigated such techniques for deploying large language models that scale efficiently on leadership-class supercomputers \cite{yin2023evaluation,dash2024optimizing}.

\subsection{Data-Parallel Training}
\begin{figure}[p]
    \centering
    \includegraphics[width=4cm]{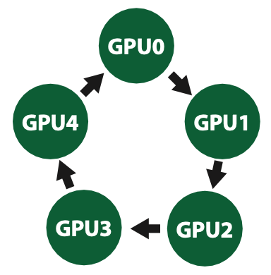}
    \caption{Ring all-reduce approach to distributed training \cite{gibiansky2017bringing}.}
    \label{fig:ring}
\end{figure}

In data-parallel training, the model is replicated across $n$ compute devices. A ``mini-batch'' of data is divided into $n$ parts, known as ``micro-batches''. Each copy of the model is trained on one micro-batch during a forward pass; the individual losses are then aggregated (typically via a ring-allreduce as shown in Fig. \ref{fig:ring}), and the gradient of the aggregated loss is back-propagated to update the model parameters in parallel.

To optimize GPU utilization, we use a large micro-batch size per device, resulting in a larger mini-batch. While large-scale high-performance computing (HPC) systems enabled data-parallel training in unprecedented scale, model convergence suffers with large mini-batches. Several approaches have been developed to mitigate these large-batch related issues: gradual warm-up, layer-wise adaptive learning rate scaling (LARS), batch-normalization, mixed-precision training, and utilization of second derivatives of loss.

\subsection{Model-Parallel Training} 

Model parallelism is typically employed when the model size or data sample size is too large to fit on a single GPU, or to improve the strong scaling of the application. The core idea of model parallelism is to place different sub-networks of a model on different devices. Compared to data-parallelism, where the model is duplicated across devices, model parallel implementations typically do not alter the parameter space of the problem, so the same hyper-parameters can work at different scales. Also, in the data-parallel method communication is limited to the backward pass; however, in most model parallel methods, communication is denser in the backward pass and is also present in the forward pass. The most straightforward method involves distributing different layers of the model among various devices: the parameters are spread across these devices, and the backward pass requires all-to-all communication. The most commonly used method is pipeline parallelism, where the layers of the model are distributed across devices without being cut, resulting in sparse communication. Pipeline parallelism is favored due to its efficient communication and is predominantly used across nodes.

In many scientific contexts, data samples are too large to fit on a single device. Spatial decomposition can be applied \cite{jin2018spatially}, wherein data is tiled across devices alongside parts of the model. This method has been successfully employed in \cite{MeshTF} for large 3D computerized tomography images. The approach can also enhance strong scaling, or time-to-solution, by distributing the compute load across devices. Efficient compute and communication overlap is essential, and the LBANN framework addresses this \cite{LBANN1, LBANN2}. Most deep learning frameworks have integrated several model parallel implementations. For instance, TensorFlow provides Mesh TensorFlow \cite{shazeer2018mesh}, Pytorch supports intranode pipeline parallelism with GPipe \cite{gpipe}, and internode parallelism via PyRPC \cite{bworld}. A comprehensive comparison is provided by \cite{bennun2018demystifying}, and efficient implementations of model parallelism across various architectures can be found in \cite{gpipe, narayanan2021efficient}. The study in \cite{LBNL} simulated the communication costs for all three types of parallelism and their combinations: the parallel strategies mentioned are largely orthogonal to each other. When model parallelism is implemented, a combination of different model methods and data-parallel is often employed, depending on the communication bottlenecks of the application \cite{narayanan2021efficient}.

\begin{figure}[hbtp] 
\begin{mdframed}[backgroundcolor=black!10,rightline=false,leftline=false]
\textbf{Example: Cancer Detection}

Much of the cancer detection research focuses on Whole Slide Images (WSI). WSIs are digital microscopy images acquired at very high pixel-level resolution. For example, a standard 4 x 6 cm glass slide at 40x magnification, after digitization turns into 200,000 x 300,000 pixels. A plethora of deep learning research has been performed around WSIs, and due to the extreme resolution of those images, the developed workloads are very well suited for HPC environments. The most common method is by patching the image online, and then training the model on a few selected patches. This method works very well in some cases; however, it has some major limitations. For example, pixel-level information is required, and careful post-processing is usually necessary due to the large number of false positives it creates \cite{Zhong2018TEAMHM}. The state-of-the-art method on WSIs uses multiple-instance learning (MIL) algorithms \cite{maron1997framework}, in which a large-size image is divided into multiple smaller images \cite{lu2021data}. The final decision is made by a weakly-supervised training model on the extracted features from the image patches \cite{das2018multiple}. There are more recent self-supervised approaches \cite{chen2022scaling, tsaris2023scaling}, although the performance and the generality of those are not as mature.
\end{mdframed}
\end{figure}

During the process of training a neural network, hyperparameter optimization (HPO) — the process of adjusting training-related hyperparameters (e.g., batch size, learning rate) and neural architectures (e.g., number of units per layer, number of layers, dropout rate) — is essential.There are different types of approaches for HPO, primarily either random search, Bayesian, or genetic algorithms. KerasTuner has recently become a popular approach for tuning TensorFlow models, which can be distributed across multiple nodes using a chief-worker strategy to allow for hundreds or thousands of differently configured neural networks to be trained in parallel. One of the novel algorithms that KerasTuner implements is HyperBand \cite{li2017hyperband}, which achieves speedup by trying many different combinations, but using a ``principled early stopping'' mechanism to discern which hyperparameters are optimal, which means the training events do not have to be run for hundreds of epochs to full completion. Ray Tune \cite{liaw2018tune} is another widely used framework for distributed hyperparameter optimization that supports Hyperband, grid search, Bayesian optimization, and population-based methods. 

Here, we list several examples which uses genetic algorithms for performing neural architecture search (NAS) on supercomputers. Genetic algorithms mimic the mutation selection process found in nature. By representing the neural network architecture as an array of bit strings (i.e., its genome), where each entry represents for example, the number of units or filters in a layer, or the dropout rate, a new candidate population can be formed by the reproduction operations (pairwise combination of two networks), then stochastically adding mutations, and then using selection to select the $N$ fittest candidate architectures, where the fitness is typically defined according as the most accurate models. Multi-node Evolutionary Neural Networks for Deep Learning (MENNDL) uses such an approach to optimally design an convolutional neural network, utilizing a large number of compute nodes \cite{young2015optimizing}. DeepHyper \cite{balaprakash2018deephyper} is an open source framework for performing NAS on HPC, which use the Balsam \cite{balsam2020distributed} framework to hide the complexities of scaling up the workflow on HPC. Such methods are able to speed up the development of designing optimal neural network architectures significantly. 

\section{AI-HPC Workflows}
\label{sec:workflows}

Over the years, several design patterns or execution motifs have emerged in solving AI problems on HPC. Brewer et al. \cite{brewer2024ai} identify six execution motifs depicting how these techniques are deployed on HPC for scientific problems. The \textit{steering} motif utilizes a trained ML model to steer an ensemble of simulations, such as \cite{ward2021colmena}. \textit{Multistage pipelining} is an abstraction typically instantiated as high-throughput virtual screening (HTVS) pipelines that are typically used in drug discovery and the design of materials \cite{woo2021optimal}. \textit{Inverse design} methods use machine learning methods to compute optimal solutions, e.g., in the search for new materials, without necessarily having to invert differential equations \cite{wang2022inverse}. \textit{Model duality} refers to hybrid AI/ML simulations \cite{partee2022using, yin2022strategies, boyer2022scalable}, also referred to as cognitive simulations (CogSim) \cite{wyatt2021disaggregation}, and such as used in digital twins \cite{pathak2022fourcastnet}. \textit{Distributed models and dynamic data} are used in federated workflows, which span across an edge-to-cloud continuum \cite{rosendo2022distributed, prigent2022supporting}. Finally, the \textit{adaptive execution} motif describes the training of large scale AI/ML models, such as large language models (LLM), using techniques such as hyperparameter optimization or neural architecture search \cite{young2015optimizing}. 

\begin{figure}[t] 
\begin{mdframed}[backgroundcolor=black!10,rightline=false,leftline=false]
\textbf{Example: Computational Steering}

Scientific discoveries are often complex and involve multiple stages of decision-making, which have traditionally been carried out by human experts. Neutron scattering is one such experimental technique that involves initial scans to identify regions of interest and refined scans to take accurate measurements of atomic and magnetic structures and dynamics of matter. This process requires a beamline scientist to review the scan images and make decisions regarding the next set of experiment parameters, which can take several days to complete. However, with the advent of AI, particularly in computer vision, there is now an opportunity to develop AI-enabled methods to \textit{steer experiments}. This has the potential to significantly reduce the time required for the decision-making process and improve the accuracy of experimental results. With the increasing computing power available today, AI-driven discoveries can be deployed at the edge through an autonomous workflow. Yin et al. \cite{yin2023toward} present an autonomous edge workflow for a neutron scattering experiment at Oak Ridge National Laboratory (ORNL), which utilizes AI to steer the experiment. This workflow has the potential to revolutionize the field of experimental science by enabling faster and more accurate decision-making, ultimately leading to more efficient and effective scientific discoveries.
\end{mdframed}
\end{figure}

\section{Conclusion}
\label{sec:conclusion}

In this paper, we explored using AI for large-scale science on supercomputers. We introduced AI4S, emphasizing its significance, distinctive data modality, methods, application domains, and workflows. We delved into the various computational methods that are used for scaling such workflows, as well as the numerical methods which may be used to solve specific types of scientific problems. We covered a wide range of examples in various areas of science, and also covered specific use cases in more detail. Looking into the future of scalable AI, we outline these considerations:

\begin{enumerate}
\item \textit{Increasingly hybrid}. While there has been a considerable amount of research in developing neural networks, which may be used in lieu of traditional simulations \cite{hennigh2021nvidia}, we do not foresee traditional simulations completely being replaced by AI anytime soon. Rather, we see more hybrid HPC-AI applications, in the form of Cognitive Simulations ``CogSim'' or Digital Twins. 

\item \textit{Mixture-of-experts over monoliths}.
While large monolithic models have generally shown better performance with the downstream scientific tasks, their training cost is prohibitively expensive. An alternative is sparsely connected mixture of experts (MoE) where it is possible to scale the number of parameters many-fold at a fractional cost of a monolithic counterpart. GPT-4 is rumored to be a mixture of eight 220B parameter models. However, this may lead to complex inference pipelines. On the HPC front, model inter-communication can become a challenge. 

\item \textit{Complex inferencing pipelines}. The introduction such methods as MoE and Hybrid AI/Sim will require more complex training/inferencing pipelines. For such systems to run efficiently on HPC, new innovative hardware that can handle the significant amounts of data movement will be important. 

\item \textit{AI for Autonomous Lab}. With the Department of Energy (DOE) prioritizing integrated research infrastructure (IRI), we foresee a pivotal role for AI, particularly foundation models, in the realization of autonomous, self-driving laboratories. These cutting-edge facilities can harness the power of computational resources to enable real-time decision-making within the experimental environment, ultimately expediting scientific discovery.

\item \textit{Resurrection of Linear RNNs}. Because transformer-based LLMs are limited by context length and computationally expensive, i.e., training speed is quadratic in length and attention requires full lookback for inference, there have been significant efforts recently looking into attention alternatives, specifically in the form of Linear RNNs \cite{rush2023, peng2023rwkv, sun2023retentive, gu2023mamba}. While linear RNNs typically do not learn as effectively as attention-based models, their main advantage lies in their greater computational efficiency, especially for long token lengths.

\item \textit{Operator-based models for solving PDEs}. The beginning of the move to operator-based models that use the technology of neural networks will make AI much more useful for simulation of PDEs. The ability to infer functions from input functions in a sampling-independent way will make models, once trained, useful to solve an entire class of related problems rather than just one.

\item \textit{The crucial role of multi-modal AI for general-purpose foundation models}. There has been extensive research in vision-language (VL) models \cite{chen2023pali,alayrac2022flamingo}, but the scale in which those models are trained is still far behind LLMs. Also, most of the largest multi-modal models target VL image-level tasks rather than VL region-level localization tasks. As large-scale multi-modal datasets become available, and unified model architecture approaches are widely adopted \cite{reed2022a,zhang2023metatransformer}, we might see at the same level of scaling multimodal models as LLMs, unlocking a much wider application potential. 

\item \textit{Importance of interpretability/explainability}. Many scientists are skeptical of AI/ML methods for science. To address such concerns, researchers have been developing tools to explain the rationale behind inference results. Class Activation Mapping (CAM)~\cite{oquab2015object} and GRADient-weighted Class Activation Mapping (Grad-CAM)~\cite{selvaraju2017grad} can highlight important regions of an image while using a CNN model. Attention map visualization is a related concept applicable to transformer based models. Bringing the interpretation techniques to modern AI models is a timely endeavour.  

\item \textit{Emergence of science-inspired and science-informed neural network architectures}. While transformer-based language models are becoming ubiquitous in various scientific applications, a new direction that maps underlying physical, chemical, and biological processes through the attention mechanism has come to the fore. Problems in biochemistry and structural molecular biology, such as protein folding (AlphaFold~\cite{alquraishi2019alphafold}) and molecular docking (TankBind~\cite{lu2022tankbind}) have benefited from such novel architectures.

\item \textit{Development of AI4S Benchmarks}. Finally, continuing to develop higher-level workflows which make training and deploying such systems will be important, as well as benchmarks that are able to assess AI4S workflow performance, as opposed to simple throughput of training or inference performance \cite{thiyagalingam2022scientific}.

\end{enumerate}

\noindent As we look to the future in a post-ChatGPT world, where AI's outperform humans in many tasks, it is clear that using such techniques will be critical and essential to continue to push the boundaries of science forward.

\section*{Acknowledgements}
This research was sponsored by and used resources of the Oak Ridge Leadership Computing Facility (OLCF), which is a DOE Office of Science User Facility at the Oak Ridge National Laboratory supported by the U.S. Department of Energy under Contract No. DE-AC05-00OR22725.\\

\section*{Declaration of generative AI and AI-assisted technologies in the writing process}
During the preparation of this work the authors used GPT-4 in order to improve grammar structures in certain places. After using this tool/service, the authors reviewed and edited the content as needed and take full responsibility for the content of the publication.

\bibliographystyle{unsrt}  
\bibliography{main}

\end{document}